\documentclass[10pt,twocolumn,letterpaper]{article}

\usepackage[pagenumbers]{cvpr} 

\usepackage[dvipsnames]{xcolor}

\definecolor{cvprblue}{rgb}{0.21,0.49,0.74}
\usepackage[pagebackref,breaklinks,colorlinks,citecolor=cvprblue]{hyperref}
\usepackage{booktabs}
\usepackage[para]{threeparttable}
\usepackage{amsmath}
\usepackage{amssymb}
\usepackage{graphicx}
\usepackage{xcolor}
\usepackage{bbm}

\usepackage{pgfplots} 
\usepgfplotslibrary{groupplots,statistics,fillbetween, external}
\pgfplotsset{compat=1.18} 

\DeclareMathOperator*{\argmin}{arg\,min}

\def\paperID{*****} 
\def\confName{IV}
\def\confYear{2024}

\title{Guiding Attention in End-to-End Driving Models
}

\author{Diego Porres$^{1}$,
        Yi Xiao$^{1}$, 
        Gabriel Villalonga$^{1}$, 
        Alexandre Levy$^{1}$, 
        Antonio M. López$^{1,2}$ 
\thanks{$^{1}$Computer Vision Center (CVC) and $^{2}$Dpt. Ciències de la Computació, Universitat Autònoma de Barcelona (UAB), Spain} 
\thanks{Corresponding author: {\tt\small diego.porres@cvc.uab.es}}%
}

\makeatletter
\def\thanks#1{\protected@xdef\@thanks{\@thanks
        \protect\footnotetext{#1}}}
\makeatother

\begin{document}
\maketitle
\begin{abstract}
Vision-based end-to-end driving models trained by imitation learning can lead to affordable solutions for autonomous driving. However, training these well-performing models usually requires a huge amount of data, while still lacking explicit and intuitive activation maps to reveal the inner workings of these models while driving. In this paper, we study how to guide the attention of these models to improve their driving quality and obtain more intuitive activation maps by adding a loss term during training using salient semantic maps. In contrast to previous work, our method does not require these salient semantic maps to be available during testing time, as well as removing the need to modify the model's architecture to which it is applied. We perform tests using perfect and noisy salient semantic maps with encouraging results in both, the latter of which is inspired by possible errors encountered with real data. Using CIL++ as a representative state-of-the-art model and the CARLA simulator with its standard benchmarks, we conduct experiments that show the effectiveness of our method in training better autonomous driving models, especially when data and computational resources are scarce.
\end{abstract}    
\section{Introduction}\label{sec:introduction}
In intricate environments, human vision adeptly focuses on goal-relevant areas, while the rest undergo a more cursory catching, or may even be ignored. This purposeful and selective cognitive process is called the Visual Attention Mechanism (VAM) \cite{Barbara:1995}. Inspired by VAM, over the past decade various attention mechanisms have been proposed to improve the performance of deep neural networks in different tasks such as image classification \cite{wang:2017, woo:2018,hu:2018, bello:2019, fu:2019}, natural language processing \cite{Vaswani:2017,bahdanau:2014,luong:2015, devlin:2018,yang:2019}, and image captioning \cite{lu:2017, rennie:2017, chen:2017}. Early works \cite{hochreiter:1997, cho:2014, luong:2015, yang:2016} proposed attention mechanisms for recurrent models, where its computation is performed sequentially along the positions of input and output. This sequential nature hinders parallelization in training samples, which can be problematic with long sequences. Differently, the Transformer model \cite{Vaswani:2017} was the first to eschew recurrence and rely entirely on a self-attention mechanism to draw global dependencies between input and output.

Accordingly, the end-to-end driving model CIL++ \cite{Xiao:2023}, a pure vision-based state-of-the-art model, includes a transformer Encoder. However, as a data-driven end-to-end model, guiding its training to focus on regions of special interest remains an unsolved problem. Moreover, as a hybrid model concatenating a CNN and a Transformer Encoder, it is also difficult to obtain clear visual activation maps that could help understand the model's driving actions.

Inspired by neuroscience, where it is stated that \textit{attention is the flexible control of limited computational resources} \cite{LindsayAttention:2020}, this paper proposes an intuitive and explicit attention-learning method to effectively guide vision-based end-to-end driving models to focus more on image content relevant to the drive. In turn, visual activation maps become more understandable. Our method is only applied at training time and does not involve modifying the underlying deep architecture of the driving model. By training CIL++ with this \textit{attention guidance learning} method and using the CARLA simulator \cite{Dosovitskiy:2017}, we provide rich ablative results to show that the proposed method improves driving performance, especially under low data regimes.

Section \ref{sec:rw} summarizes the most related literature. Section \ref{sec:method} draws the Attention Guidance Learning method we propose, and Section \ref{sec:experiments} shows its effectiveness experimentally. Finally, Section \ref{sec:conclusions} summarizes the main conclusions and points toward future work.

\begin{figure*}[thpb]
 \centering \includegraphics[width=\linewidth]{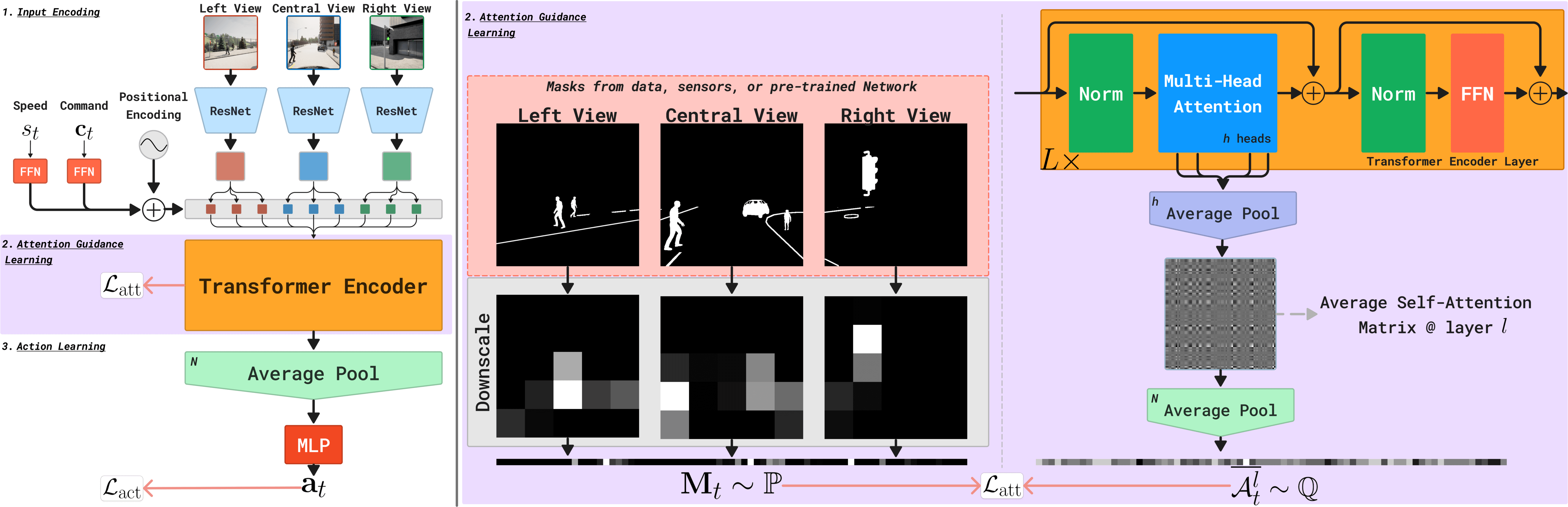}
 \caption{Our proposed pipeline. Left: the CIL++ \cite{Xiao:2023} architecture. Right: our proposed Attention Loss $\mathcal{L}_{\text{att}}$ obtained from masks using pre-computed data, on-board sensors, or a pre-trained network. For additional details, refer to Section \ref{subsec:architecture}  and \ref{subsec:loss_func}, respectively.}
 \label{fig:architecture}
\end{figure*}
\section{Related Work}
\label{sec:rw}

\subsection{Imitation Learning}
As a promising approach to training end-to-end systems, Imitation Learning (IL) has been applied to a variety of tasks, including robot manipulation \cite{Ratliff:2007, Stepputtis:2020, kim:2023}, autonomous driving \cite{George:2018, Pan:2018, Codevilla:2018, Hu:2022, Xiao:2023}, game playing \cite{jorge:2016, barratt:2019, thammineni:2023}, and piloting an aircraft \cite{Sammut:1992, freitas:2023}. These works have shown that IL is a training method that deserves further research. 

As the IL approach learns action policies directly from expert demonstrations, {\ie}, neither relying on rule-based predefined policies nor cumbersome manual data annotation, it has been a compelling research topic in autonomous driving \cite{Codevilla:2018, Pan:2018, Liang:2018, Bewley:2019, Codevilla:2019, Xiao:2020, Zhang:2021, Hu:2022, Xiao:2023, jamgochian:2023}. These pioneering works illustrate the possibility of mapping sensor data straight to the vehicle's control signals (steering angle and brake/throttle) through the use of deep neural networks, without the need for intermediate modules such as semantic perception and local path planning.

As with any data-driven method, IL needs to address the dataset bias problem and causal confusion \cite{Codevilla:2019, chen:2023}. Moreover, understanding the causality between the input and the output is difficult since no explicit intermediate semantic representation is available.

\subsection{Attention Guidance Learning}
The human visual system makes use of attention mechanisms to facilitate efficient processing. Indeed, human eyes capture redundant visual inputs that the brain can naturally process to highlight only relevant information for the targeted goal. This encourages including similar mechanisms to develop deep learning models. For instance, this idea has already brought benefits in many Computer Vision tasks \cite{wang:2017, woo:2018,hu:2018, bello:2019, fu:2019, lu:2017, rennie:2017, chen:2017, Alexey:2021, guo:2022}. Including attention in Computer Vision can be traced back to 2014, when Mnih {\etal} \cite{mnih:2014} presented an RNN model that is capable of extracting information from an image or video by an adaptive selection of a sequence of regions that are then only processed at high resolution. Jaderberg {\etal} \cite{jaderberg:2015} introduced a learnable module, the Spatial Transformer, that allows networks to not only select regions of an image that are most relevant but also to transform them into a canonical simplified representation. More recently, along with the proposals of Transformer models such as BERT \cite{devlin:2018} and ViT \cite{Alexey:2021}, the idea of self-attention \cite{Vaswani:2017} has rapidly attracted great interest. Various Transformer-based variants such as XLNet \cite{yang:2019}, PCT \cite{guo:2021}, Swin-Transformer \cite{liu:2021}, and Transfuser \cite{chitta:2022} have shown that attention-based models have the potential to be a powerful and general architecture in Computer Vision.

Broadly speaking, these attention mechanisms are generic and learn the specificity of the model tasks by training with enough, diverse, and representative data, which is not always available. Therefore, for vision-based driving-related tasks, different works have been proposed to explicitly force attention on specific image regions instead of learning them from scratch. The idea is to predict a saliency map that is used as an additional input channel to the RGB ones \cite{Zhang:2021attention, Ding:2023} or is used to weight the RGB channels \cite{Faisal:2021}. Sometimes these saliency maps are computed as a prediction of where human drivers would be gazing at \cite{Xia:2020}. A different approach is to use such saliency maps not as an input to the model under training but to add a loss factor to guide attention \cite{Cultrera:2023}. This last approach is conceptually aligned and potentially complementary to others that use some auxiliary perceptual tasks ({\eg}, depth estimation \cite{Ishihara:2021}, semantic segmentation \cite{Wang:2019, Chitta:2021, Ishihara:2021, Hu:2022}, object detection \cite{Wang:2019}), aiming at improving driving performance in end-to-end driving models.  

\subsection{Our Method in Context}
In this paper, we are interested in pure vision-based end-to-end driving models trained by imitation learning. These can lead to very affordable solutions for autonomous driving as there is no need for extra sensors or costly data annotation; however, they still require more research for reliability and explainability. For this reason, we will focus on the current state-of-the-art pure vision-based end-to-end driving model CIL++ \cite{Xiao:2023}, but we believe that the proposed method can be applied even to non-pure-vision models in the future. All of our experiments will run on the CARLA simulator \cite{Dosovitskiy:2017}.

We assess how to force attention at training time through pre-computed saliency maps, which we hypothesize can turn into better driving performance. In contrast to \cite{makrigiorgos2019human, Zhang:2021attention, Ding:2023}, we neither force additional input channels nor perform mask-based input-image weighting, which allows us to avoid predicting these masks during driving. 
Unlike Cultrera \etal  \cite{Cultrera:2023}, we exploit the self-attention maps in the architecture to highlight regions of interest without a need to \textit{select} regions in the image and discard the rest.

The proposed saliency maps consist of binary masks highlighting task-specific classes of interest: vehicles, pedestrians, traffic signs, lane marks, and road borders. However, as these saliency maps are available at training time, and we want to keep a setting where no manual labeling of images is performed, in practice we can assume that such masks can be provided by synth-to-real unsupervised domain adaptation (StR UDA) models \cite{gomez2023cotraining, hoyer2022MIC}. Nonetheless, we can assume these predicted masks to be noisy, and we will show our method is robust enough even with noisy masks.

Furthermore, in our experience, providing visual activation maps to help interpret CIL++ actions has been unsuccessful. In other words, even if the model drives perfectly, the obtained activation maps are far from human-readable. Applying our method to guide attention can produce intuitive activation maps, thus, opening the door for reintroducing interpretability in end-to-end driving models.

\section{Attention Learning for End-to-End Driving}
\label{sec:method}

\subsection{Problem Setup} 

Our model is trained via IL, where an expert driver provides a set of driving demonstrations that can be imitated by an agent in an end-to-end manner. Following an optimal driving policy $\pi^{\star}$ that maps each instance of observations to the available action space, the expert driver performs actions $\mathbf{a}_i$ based on a set of observations $\mathbf{o}_i$ of the current environment, {\ie}, $\pi^{\star}(\mathbf{o}_i) = \mathbf{a}_i$.
Thus, to effectively train an agent, we use the expert driver to collect a dataset comprised of observation/action pairs $\mathcal{D} = \left\{(\mathbf{o}_{i}, \mathbf{a}_{i})\right\}_{i=1}^T$. The agent will follow a policy $\pi_{\theta}$ that approximates the expert policy $\pi^{\star}$ via the general imitation learning objective

\begin{equation} \label{eq:il_target}
  \pi_{\theta} = \argmin_{\theta} \mathbb{E}_{(\mathbf{o}_{i}, \mathbf{a}_{i})\sim \mathcal{D}} \left[ \mathcal{L}(\pi_{\theta}(\mathbf{o}_{i}), \mathbf{a}_{i}) \right] \enspace .
\end{equation}

At testing time, only the trained policy  $\pi_{\theta}(\mathbf{o}_{i})$ will be used to drive the agent. 

\subsection{Architecture}\label{subsec:architecture}

Fig. \ref{fig:architecture} illustrates the overall architecture of our proposed model. Broadly, we can lump our model in the following three phases: Input Encoding, Attention Guidance Learning, and Action Learning.

\subsubsection{Input Encoding}
We keep the same setting as in CIL++ \cite{Xiao:2023}. Concretely,  at each timestep $t$ the input to the network consists of three parts: 1) a set of $K$ RGB images of dimensions $W\times H$ from the $K$ cameras $\mathbf{X}_{t}=\{\mathbf{x}_{1,t}, \dotsc, \mathbf{x}_{K,t} \}$, 2) the forward speed of the ego vehicle $s_t\in\mathbb{R}$, and 3) a high-level (one-hot) navigation command $\mathbf{c}_t$ that indicates which of the $M$ commands the ego vehicle should follow. 

We encode each of the $K$ RGB images via a shared-weight ResNet backbone pre-trained on ImageNet \cite{He2015DeepRL, Deng:2009}. The embedded output for each view is a set of feature maps from the last convolutional layer of ResNet, thus with a shape of ${w\times h\times c}$ ($c$ indicating the feature dimension). We flatten these feature maps along the spatial dimensions and concatenate them, resulting in $N=K\cdot w\cdot h$ tokens of dimension $c$. In addition, we linearly project the forward ego vehicle speed and the high-level navigation command to this same dimension $c$ via separate fully connected layers, and add them to the token sequence. To provide the positional information for each token, we add a learnable positional embedding to the entire sequence. In practice, we use $K=3$ RGB cameras (left, central, and right cameras) each with dimensions $W=H=300$ pixels, and set $c=512$ to match the dimensionality of the last ResNet block.

\subsubsection{Attention Guidance Learning} 
CIL++ \cite{Xiao:2023} adopts the self-attention mechanism of a Transformer Encoder block to associate relevant information across the $K=3$ views. We keep the same setting, leveraging the self-attention layers to associate features across views. As shown in Fig. \ref{fig:architecture}, our Transformer Encoder block consists of $L=4$ multi-head attention layers. Each one includes a Multi-headed Self-Attention (MHSA) \cite{Vaswani:2017} block with $h=4$ heads, layer normalization (LN) \cite{Ba:2016}, and feed-forward MLP blocks (FFN). The hidden dimension $D$ of the Transformer Encoder is set equal to the ResNet output dimension, \ie, $D=c=512$. Unlike the vanilla CIL++ training scheme, we select one of these $L$ layers to apply the Attention Loss to, which we further expand in Section \ref{subsubsec:att_loss}.

\subsubsection{Action Learning}
The output of the Transformer Encoder is the same shape as the input sequence, namely $N\times c$. We apply a global average pooling (GAP) along the sequence dimension $N$, obtaining a vector of dimension $1\times c$. This vector is fed into an MLP consisting of two fully connected layers with ReLu non-linearity. The final output action $\hat{\mathbf{a}}_t=(\hat{a}_{\text{s},t}, \hat{a}_{\text{acc}, t})$ comprises of the steering angle and acceleration, the latter being the difference between throttle and brake \cite{Zhang:2021,Xiao:2023}. We normalize both actions to lie in the range of $[-1, 1]$, where negative values correspond to turning left or braking, and positive values correspond to turning right or accelerating, respectively for $\hat{a}_{\text{s}, t}$ and $\hat{a}_{\text{acc}, t}$.

\subsection{Loss Function}\label{subsec:loss_func}
The total loss function is weighted by two parts: the \textit{Action Loss} and the \textit{Attention Loss}:

\begin{equation}\label{eq:total_loss}
    \mathcal{L}= \lambda_{\text{act}}\mathcal{L}_{\text{act}} + \lambda_{\text{att}}\mathcal{L}_{\text{att}}
\end{equation}
where $\lambda_{\text{act}},\lambda_{\text{att}}\in\mathbb{R}^+$ indicate the weight given to the Action and Attention Loss, respectively, which we explain in the following two phases. 

\subsubsection{Action Loss}
At each timestep $t$, given a predicted action $\hat{\mathbf{a}}_{t}\in\mathbb{R}^2$ by our network and a ground truth action $\mathbf{a}_{t}\in\mathbb{R}^2$ by the expert driver, we define the \textit{Action Loss} as:

\begin{equation}\label{eq:action-loss}
\mathcal{L}_{\text{act}}(\mathbf{a}_t, \hat{\mathbf{a}}_t) = \lambda_{\text{s}} \lVert \hat{a}_{\text{s},t}-a_{\text{s},t} \rVert_{1} + \lambda_{\text{acc}}\lVert \hat{a}_{\text{acc},t}-a_{\text{acc,t}}\rVert_{1} \enspace ,
\end{equation}

\noindent where $\lVert\cdot\rVert_{1}$ is the $L_1$ distance and $\lambda_{\text{s}},\lambda_\text{acc}\in\mathbb{R}^{+}$ indicate the weights given to the steering angle and acceleration parts, respectively. 

\subsubsection{Attention Loss}\label{subsubsec:att_loss}

We add an attention-learning branch to prompt an end-to-end driving model to intentionally heed safety-critical regions in the input images. At each timestep $t$ and for each camera $i$, these regions will be defined via a single-channel synthetic attention mask $\mathcal{M}_{i, t} \in \mathbb{R}^{W\times H}$.

To render these attention masks used as ground truth, we make assumptions about the focus of a regular driver while driving. Specifically, we make use of the semantic segmentation and depth images provided by the CARLA simulator. These features enable us to precisely isolate and emphasize specific objects or areas within the visual field of the driving simulation. Our attention masks highlight safety-critical dynamic objects such as cars and pedestrians, and static objects that are essential for navigation and driving decisions such as traffic lights, road signs, lane markings, and road borders. These are crucial indicators of the physical space within which the car can maneuver.
Furthermore, we incorporate a depth threshold in our attention mask algorithm to ensure that the driver's attention is realistically focused on elements within a practical and safe range of the ego vehicle.

We hypothesize that the attention maps of the Transformer Encoder can effectively approximate the distribution of the attention masks.  
To achieve this, we first downscale the $K$ masks to $w \times h$, matching the spatial resolution in the Input Encoding phase. We flatten, concatenate them, then normalize this sequence, resulting in a mask $\mathbf{M}_t\in \mathbb{R}^N$ that follows the target distribution, \ie, $\mathbf{M}_t \sim \mathbb{P}$.
We take advantage of the distributional property of the self-attention maps of the Transformer Encoder and force them to match $\mathbb{P}$. In practice, we select a layer $l$, get the average attention matrix of the $h$ heads, average it row-wise $\overline{\mathcal{A}_{t}^{l}} = \mathbb{E}_N[ \mathbb{E}_h[\mathcal{A}_{t}^{h, l}]]\in\mathbb{R}^N$, and then compare both distributions using the Kullback-Leibler (KL) divergence as our \textit{Attention Loss}. We can reformulate it pointwise  (and time-step-wise) as:

\begin{equation}\label{eqn:attn_loss}
\mathcal{L}_{\text{att}}(\mathbf{M}_{t}, \overline{\mathcal{A}_{t}^{l}}) = \sum_{j=1}^{N}  \mathbf{M}_{t}^j  \log \left({\epsilon+\frac{\mathbf{M}_{t}^j} {\overline{\mathcal{A}_{t}^{l}}^j + \epsilon}} \right)
\end{equation}

where $\epsilon$ is a small number added for regularization. In practice, we apply this loss at the last layer $l=L=4$, but there is no limitation on which layer to apply it to, nor to which heads. We leave the latter for future work.

\paragraph{Realistic Masks}\label{p:real_masks}We define a function $f(\mathcal{M}_{i, t})$ to introduce realistic noise into masks $\mathcal{M}_{i, t}$ using depth-aware Perlin noise \cite{perlin:2002}, creating variable intensity `blobs' that mimic real-world imperfections. It is refined so that the distortions extend beyond simple blobs, introducing more granular disturbances on larger objects like cars and red lights. It deliberately excludes thinner features such as lane markings, which are too subtle for detailed granularity. This enhancement more accurately simulates typical real-world noise artifacts that arise from sensor or cumulative errors by StR UDA models \cite{gomez2023cotraining, hoyer2022MIC}. Fig. \ref{fig:masks} showcases examples of our noise-integrated masks.

\section{Experiments}
\label{sec:experiments}

\subsection{Driving Environments }
We conduct our experiments on the CARLA simulator \cite{Dosovitskiy:2017}, version \texttt{0.9.13}. Regarding the expert driver used for data collection, we use a \textit{teacher} model (agent) from \cite{Zhang:2021} which is based on reinforcement learning and thus shows a more realistic and diverse behavior than the default expert driver in CARLA. We keep the same settings of CIL++ \cite{Xiao:2023}, using three forward-facing onboard cameras that cover a horizontal field-of-view of $180^{\circ}$ in total ($60^{\circ}$ for each camera without overlapping). For the input RGB images, the resolution is set to $W\times H=300\times300$ pixels.

To ensure consistency and reliability in our assessment, we used the same agent for all the training datasets. Data collection occurred at a rate of 10 FPS, given a spectrum of weather conditions and towns within the CARLA environment. The resulting datasets, namely the 14 and 55-hour datasets, were generated to validate and show the efficacy of our approach under varying weather conditions.

The \textbf{14-hour dataset}, acquired in \texttt{Town01}, featured eight hours of driving with a "busy" object density as specified in \cite{Zhang:2021}. This dataset further diversified its scenarios by allocating two hours to each of the ClearNoon, ClearSunset, HardRainNoon, and WetNoon weather conditions. An additional six hours were recorded under the same weather conditions but with an empty object map. As \texttt{Town01} is built with single-lane roads, there are $M=4$ vehicle commands in this dataset: \textit{turning left, turning right, continue straight,} and \textit{follow the lane.}

In contrast, the \textbf{55-hour dataset} was designed to explore the adaptability of our approach in a more complex environment, spanning \texttt{Town01} to \texttt{Town06}. This dataset contains diverse driving scenarios, including multi-lane driving, highways, and crossroads. Consequently, the command set for this dataset expands to $M=6$, incorporating the commands defined in the 14-hour dataset and introducing lane-change directives: \textit{change to left lane} and \textit{change to right lane}.

\begin{figure*}
\centering
\setlength{\tabcolsep}{1pt} 

\renewcommand*{\arraystretch}{0}
\begin{tabular}{ c c c c c c}
   & $\mathbf{x}_{c,t}$ & $\mathcal{M}_{c, t}$ & $f(\mathcal{M}_{c,t})$ &  $\widehat{\mathcal{M}}_{c, t}$ &  $f(\widehat{\mathcal{M}}_{c, t})$ \\
  \rotatebox{90}{\texttt{Town01}} & \includegraphics[width=0.19\textwidth]{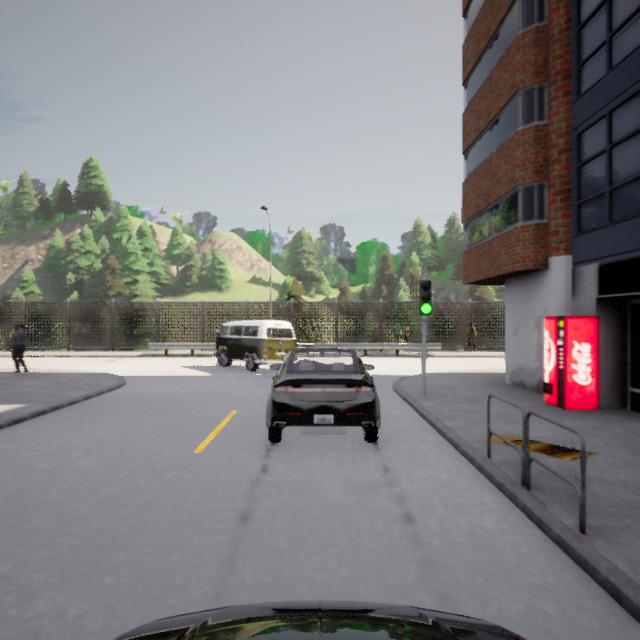} & 
    \includegraphics[width=0.19\textwidth]{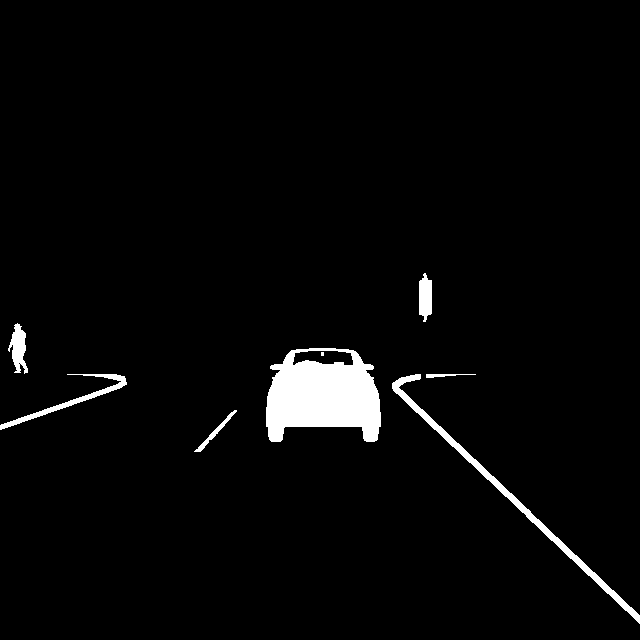} & 
    \includegraphics[width=0.19\textwidth]{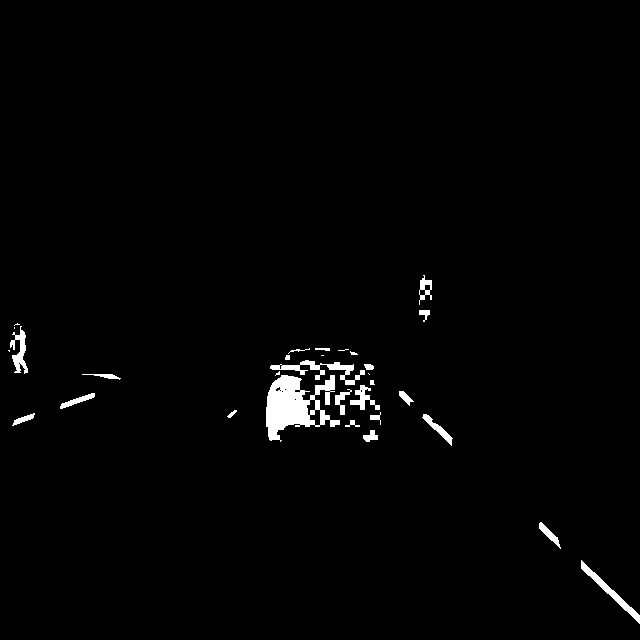} & 
    \includegraphics[width=0.19\textwidth]{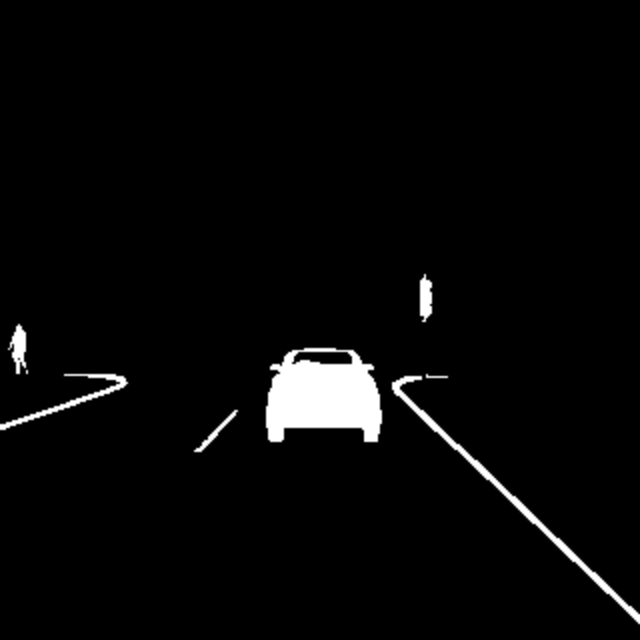} & 
    \includegraphics[width=0.19\textwidth]{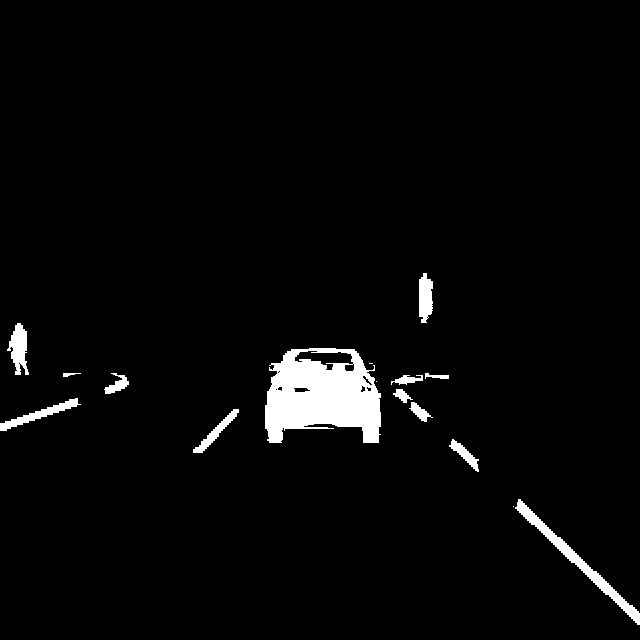} \\
  \rotatebox{90}{\texttt{Town02}} & \includegraphics[width=0.19\textwidth]{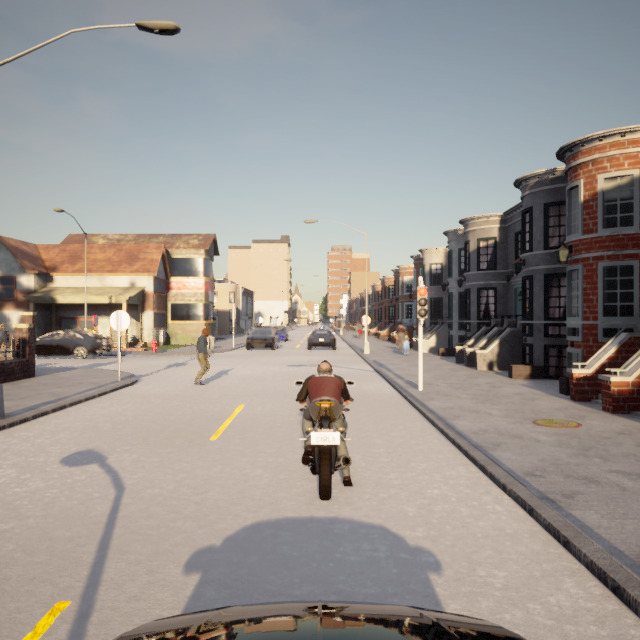} & 
    \includegraphics[width=0.19\textwidth]{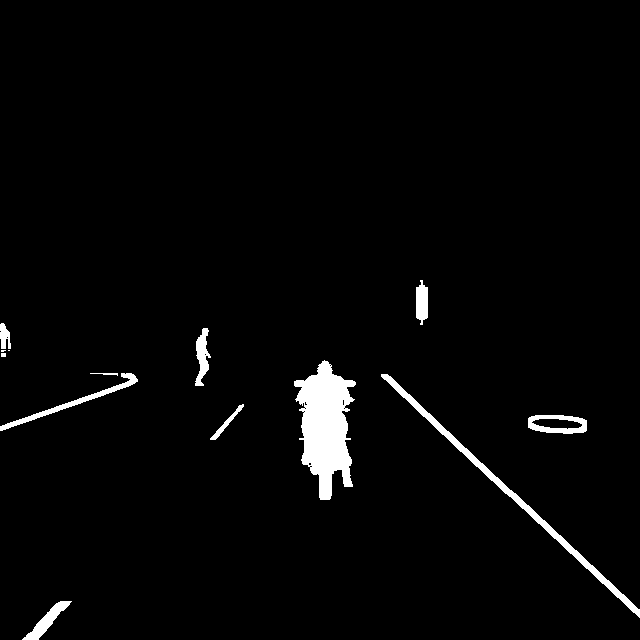} & 
    \includegraphics[width=0.19\textwidth]{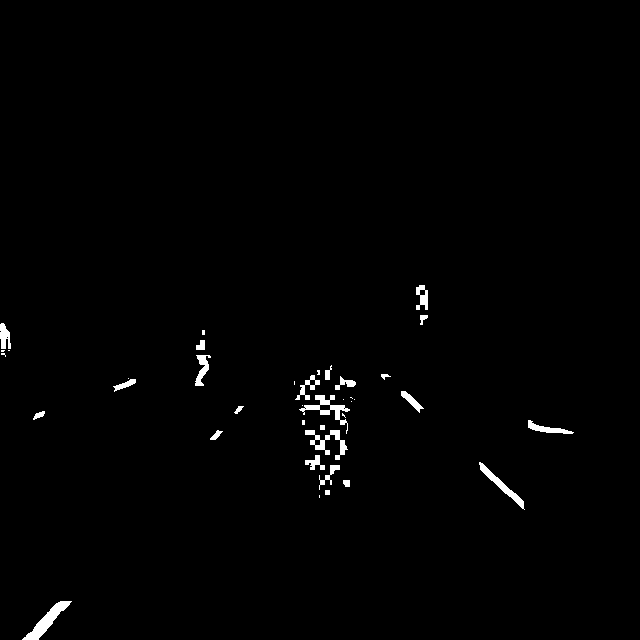} & 
    \includegraphics[width=0.19\textwidth]{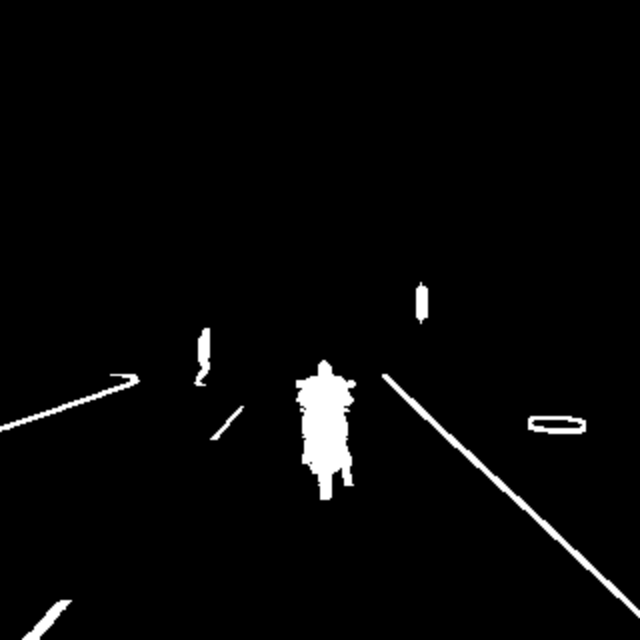} & 
    \includegraphics[width=0.19\textwidth]{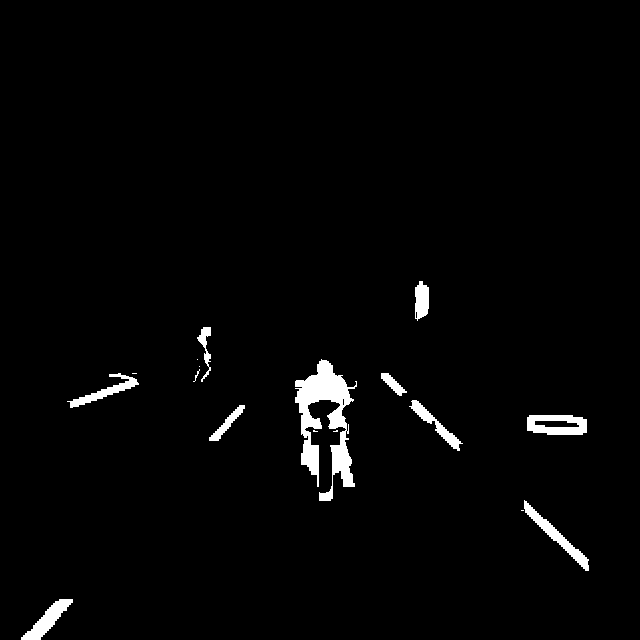} \\
  \rotatebox{90}{\texttt{Town03}} & \includegraphics[width=0.19\textwidth]{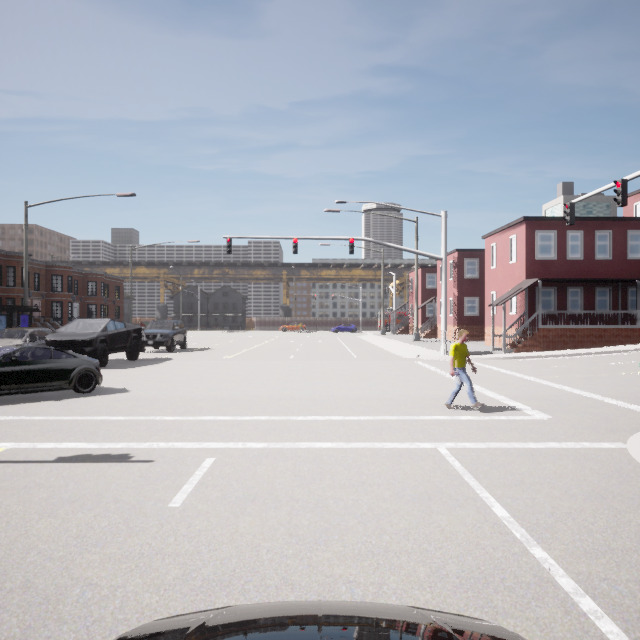} & 
    \includegraphics[width=0.19\textwidth]{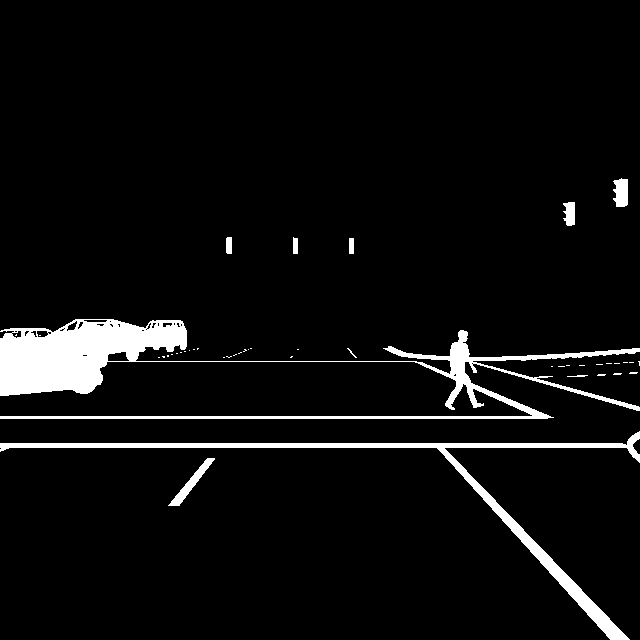} & 
    \includegraphics[width=0.19\textwidth]{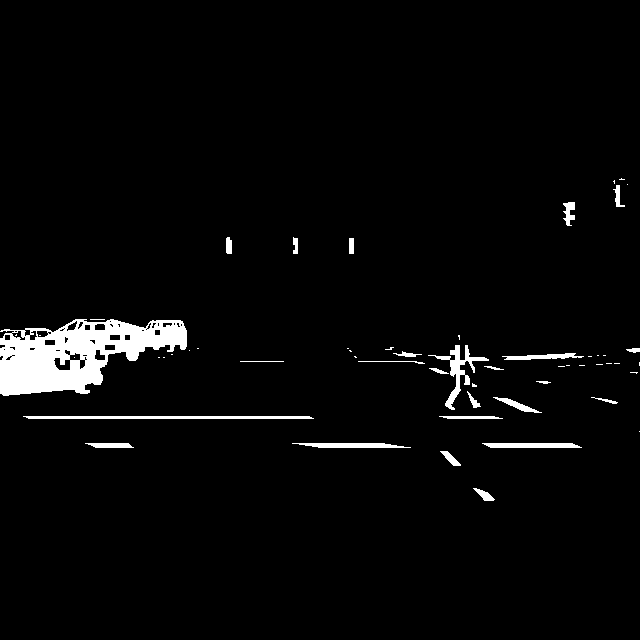} & 
    \includegraphics[width=0.19\textwidth]{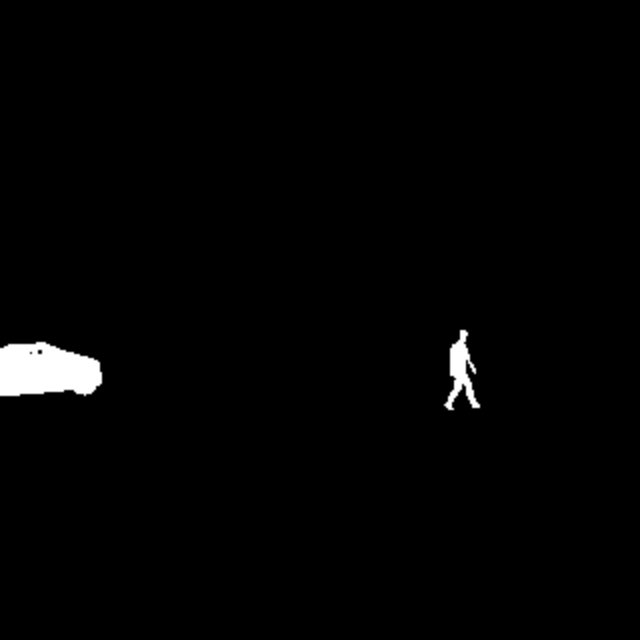} & 
    \includegraphics[width=0.19\textwidth]{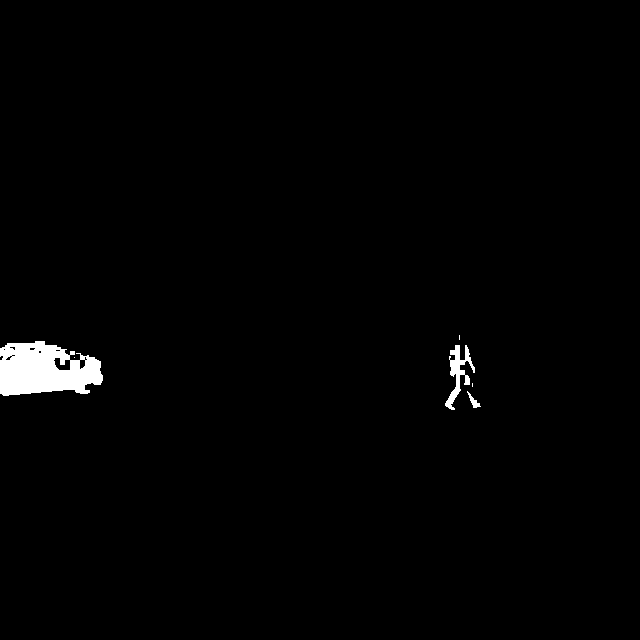} \\
\end{tabular}
\caption{$\mathbf{x}_{c, t}$ (central RGB images at a timestep $t$) and their corresponding masks $\mathcal{M}_{c, t}$ for \texttt{Town01}, \texttt{Town02}, and \texttt{Town03}. For the single-lane (top rows), we use a maximum depth of 20 meters to generate the masks, whereas we use a maximum depth of 40 meters for the multi-lane towns. Note that the U$^2$-NET was trained only with data from \texttt{Town01}, so the failure to detect the lanes on \texttt{Town03} is merely illustrative.}
    \label{fig:masks}
\end{figure*}

\subsection{Validation}
We will validate the trained models in an online manner (using dynamic agents), so we will not use a static dataset. This evaluation is done by driving in specific scenarios and evaluating the driving quality of the agent. During this, we set the CARLA simulator in a \texttt{synchronous} manner, that is, the model can be validated without being affected by its inference time.

When training with the 14-hour dataset and its smaller subsets, we validate in the unseen \texttt{Town02} from CARLA as it also contains single-lane roads. We use the \texttt{NoCrash} setup \cite{Codevilla:2019} with two different weather conditions than those seen during training. For the 55-hour dataset, we use the offline Leaderboard from CARLA, using 10 distinct routes in \texttt{Town05} under two new validation weather configurations. In both, the new weather configurations are SoftRainSunset and WetSunset.

In the NoCrash setup, the ego agent continues to navigate until reaching the end of the route, unless it collides with some object or a time-out happens, irrespective of other driving infractions. This is not the case for the offline Leaderboard where crashes may occur. For both, we report the following key metrics extracted from the CARLA Leaderboard benchmark:

\begin{itemize}
    \item \textbf{Success Rate (SR)} indicates the percentage of routes where the car successfully reaches the destination. It serves as a measure of the agent's ability to complete the designated routes.
    \item \textbf{Route Completion (RC)} is the average of the route the ego vehicle managed to accomplish (as a percent) for all routes.
    \item \textbf{Infraction Score (IS)} is a scoring metric that quantifies the number of driving infractions on each route, which include collisions with objects, ignoring traffic lights and stop signals and other rule violations. "No infractions" is indicated as 1, decreasing with every infraction.
    \item \textbf{Driving Score (DS)} is the product of the RC and IS per route. This is a combined metric that considers all aspects of a driving agent.
\end{itemize}

We repeat our driving test three times with different random seeds, as we randomly spawn and control the other vehicles and pedestrians in the simulator. This way, we aim to offer a detailed and quantitative assessment of our agent's performance in diverse driving scenarios.

\subsection{Training Hyperparameters}
In our experiments, the CIL++\cite{Xiao:2023} public code serves as the framework for executing our trials. Regardless of the dataset employed for training, we adopt the hyperparameter settings outlined in \cite{Xiao:2023} to ensure consistency in our model training. Throughout all experiments, we keep the Action Loss weights the same, \ie, $\lambda_{\text{act}} = 1$ and  $\lambda_{\text{s}}=\lambda_{\text{acc}}=0.5$. The training process spans 80 epochs with a batch size of $512$ and initial learning rate of $10^{-4}$, and the best-performing model, determined during training, is selected for evaluation.

To optimize the performance of our models, we systematically explore different values for the Attention Loss weight, using a small dataset comprising all the collected "busy" data in \texttt{Town01} (consisting of 8 hours of driving): $\lambda_{\text{att}}\in\{0.1,0.25,0.5,1.0,2.5,5.0,10.0 \}$. We validate these models in \texttt{Town01} under different weather conditions to those seen during training. Our experiments show that a higher weight in Attention Loss significantly enhances results, showing a notable improvement of up to 47 points in the Success Rate metric compared to cases where attention loss is not applied. Based on these findings, we adopt a value of $\lambda_{\text{att}}=10$.

\subsection{Quantitative Results}

In our experimental design, we categorize the results into two distinct sections: the \textit{low data regime} and the \textit{high data regime}. The low data regime is specifically dedicated to evaluating the performance of our approach when confronted with a lack of data along two axes: low availability (number of driving hours) and low variability (decreasing the types of weather conditions in the dataset). 

On the other hand, the high data regime is strategically designed to compare results with robust datasets and various baselines relevant to our work. This division of our experimental results facilitates an analysis of the adaptability, robustness, and comparative performance of our approach across different data availability scenarios.

\subsubsection{Low data regime}
To assess the impact of training with Attention Loss, we conduct experiments using incremental subsets of the 14-hour dataset with a relatively busy traffic density, randomly sampled starting from 2 hours and increasing in 2-hour increments up to 8 hours. Additionally, we include results obtained with the full 14-hour dataset for a comprehensive comparison. 

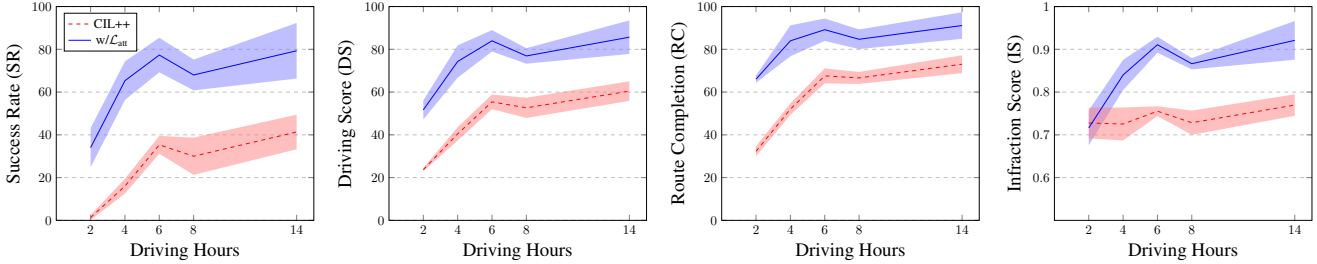
\begin{figure*}[ht]
\centering
\resizebox{\linewidth}{!}{\begin{tikzpicture}
\begin{groupplot}[
    group style={
        group size=4 by 1,
        vertical sep=2cm,
        horizontal sep=2cm
    },
    xmin=0, xmax=15,
    ymin=0, ymax=100,
    ymajorgrids=true,
    grid style=dashed,
    legend pos=north west,
]

\nextgroupplot[
    xlabel={\Large Driving Hours},
    ylabel={\Large Success Rate (SR)},
    xtick={2,4,6,8,14},
    legend entries={w/$\mathcal{L}_{\text{att}}$, CIL++}
]
    \addplot[color=red, dashed] coordinates {(2,1.33)(4,16)(6,35.33)(8,30)(14,41.33)};\addlegendentry{CIL++}
    \addplot[color=blue] coordinates {
    (2,34)(4,65.33)(6,77.33)(8,68)(14,79.33)}; \addlegendentry{w/$\mathcal{L}_{\text{att}}$}
    \addplot[name path=sr_max,color=blue!70,draw=none] coordinates {(2,43.17)(4,74.35)(6,85.42)(8,75.21)(14,92.35)};
    \addplot[name path=sr_min,color=blue!70,draw=none] coordinates {(2,24.83)(4,56.31)(6,69.25)(8,60.79)(14,66.32)};
    \addplot[blue!50,fill opacity=0.5] fill between[of=sr_max and sr_min];
    \addplot[name path=cpp_max,color=red!70,draw=none] coordinates {(2,2.49)(4,19.46)(6,39.5)(8,38.72)(14,49.42)};
    \addplot[name path=cpp_min,color=red!70,draw=none] coordinates {(2,0.18)(4,12.54)(6,31.17)(8,21.28)(14,33.25)};
    \addplot[red!50,fill opacity=0.5] fill between[of=cpp_max and cpp_min];

\nextgroupplot[
    xlabel={\Large Driving Hours},
    ylabel={\Large Driving Score (DS)},
    xtick={2,4,6,8,14},
]
    \addplot[color=red,dashed] coordinates {(2,23.73)(4,40.4533)(6,55.425)(8,52.637)(14,60.45)};
    \addplot[color=blue] coordinates {
    (2,51.694)(4,74.295)(6,83.9777)(8,76.903)(14,85.6747)};
    \addplot[draw=none, name path=sr_max,color=blue!70] coordinates {(2,56.1886)(4,81.8609)(6,88.9917)(8,80.5362)(14,93.5186)};
    \addplot[draw=none, name path=sr_min,color=blue!70] coordinates {(2,47.1994)(4,66.7291)(6,78.9636)(8,73.2698)(14,77.8307)};
    \addplot[blue!50,fill opacity=0.5] fill between[of=sr_max and sr_min];
    \addplot[draw=none, name path=cpp_max,color=red!70] coordinates {(2,24.5052)(4,43.5759)(6,58.8506)(8,57.3535)(14,65.0506)};
    \addplot[draw=none, name path=cpp_min,color=red!70] coordinates {(2,22.9548)(4,37.3308)(6,51.9994)(8,47.9205)(14,55.8494)};
    \addplot[red!50,fill opacity=0.5] fill between[of=cpp_max and cpp_min];

\nextgroupplot[
    xlabel={\Large Driving Hours},
    ylabel={\Large Route Completion (RC)},
    xtick={2,4,6,8,14}
]

    \addplot[color=red,dashed] coordinates {(2,32.3207)(4,52.024)(6,67.5827)(8,66.6167)(14,73.033)};
    \addplot[color=blue] coordinates {
    (2,66.0583)(4,83.9703)(6,89.1713)(8,84.6963)(14,91.129)}; 
    \addplot[draw=none, name path=sr_max,color=blue!70] coordinates {(2,67.774)(4,91.2028)(6,94.4171)(8,89.2861)(14,97.3361)};
    \addplot[draw=none, name path=sr_min,color=blue!70] coordinates {(2,64.3427)(4,76.7378)(6,83.9255)(8,80.1065)(14,84.9219)};
    \addplot[blue!50,fill opacity=0.5] fill between[of=sr_max and sr_min];
    \addplot[draw=none, name path=cpp_max,color=red!70] coordinates {(2,34.6803)(4,54.7209)(6,71.0906)(8,69.4192)(14,77.2117)};
    \addplot[draw=none, name path=cpp_min,color=red!70] coordinates {(2,29.9611)(4,49.3271)(6,64.0747)(8,63.8142)(14,68.8543)};
    \addplot[red!50,fill opacity=0.5] fill between[of=cpp_max and cpp_min];

\nextgroupplot[
    xlabel={\Large Driving Hours},
    ylabel={\Large Infraction Score (IS)},
    xtick={2,4,6,8,14},
    ymin=0.5,ymax=1,
    ytick={0.6,0.7,0.8,0.9,1.0},
]

    \addplot[color=red, dashed] coordinates {(2,0.7277)(4,0.7253)(6,0.7553)(8,0.7283)(14,0.7697)};
    \addplot[color=blue] coordinates {
    (2,0.7163)(4,0.84)(6,0.9107)(8,0.8663)(14,0.9210)}; 
    \addplot[draw=none, name path=sr_max,color=blue!70] coordinates {(2,0.7572)(4,0.8753)(6,0.9294)(8,0.8796)(14,0.9663)};
    \addplot[draw=none, name path=sr_min,color=blue!70] coordinates {(2,0.6755)(4,0.8047)(6,0.8919)(8,0.8530)(14,0.8757)};
    
    \addplot[blue!50,fill opacity=0.5] fill between[of=sr_max and sr_min];
    \addplot[draw=none, name path=cpp_max,color=red!70] coordinates {(2,0.7638)(4,0.7639)(6,0.7668)(8,0.7567)(14,0.7948)};
    \addplot[draw=none, name path=cpp_min,color=red!70] coordinates {(2,0.6915)(4,0.6868)(6,0.7438)(8,0.7000)(14,0.7445)};
    
    \addplot[red!50,fill opacity=0.5] fill between[of=cpp_max and cpp_min];
    
\end{groupplot}
\end{tikzpicture}}
\caption{Comparison between the baseline (CIL++ default training) and our method (with $\mathcal{L}_{\text{att}}$) while increasing the amount of training data.}
\label{fig:lowdata_town02nocrash_meanstdev}
\end{figure*}

The results, presented in Fig. \ref{fig:lowdata_town02nocrash_meanstdev}, highlight the performance improvement when employing Attention Loss. When examining the baseline results for the 2-hour and 4-hour subsets, it becomes evident that the limited data availability hinders the driver's performance, reflected in a lower SR metric of only 0 and 16 points respectively, half the performance with more extensive datasets. Conversely, when Attention Loss is applied, the dependency on large amounts of data diminishes. Even with only 4 hours of data, the model achieves a notable average SR metric of 65, representing a significant improvement of 49 points compared to the case when the model is trained without Attention Loss. Furthermore, when we add data using Attention Loss, the IS metric shows a clear improvement in driving quality.

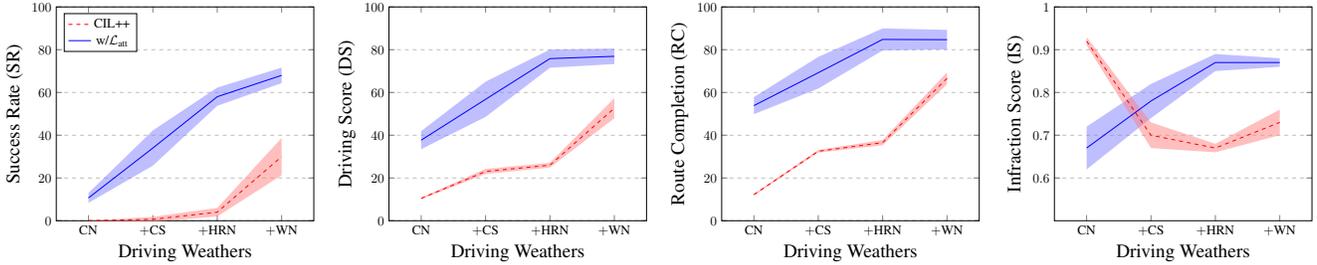
\begin{figure*}[ht]
\centering
\resizebox{\linewidth}{!}{\begin{tikzpicture}
\begin{groupplot}[
    group style={
        group size=4 by 1,
        vertical sep=2cm,
        horizontal sep=2cm
    },
    xmin=0.5, xmax=4.5,
    ymin=0, ymax=100,
    ymajorgrids=true,
    grid style=dashed,
    legend pos=north west,
    xticklabels={%
        CN,%
        $+$CS,%
        $+$HRN,%
        $+$WN,%
    },
]

\nextgroupplot[
    xlabel={\Large Driving Weathers},
    ylabel={\Large Success Rate (SR)},
    xtick={1,2,3,4},
    legend entries={w/$\mathcal{L}_{\text{att}}$, CIL++}
]
    \addplot[color=red,dashed] coordinates {(1,0.0)(2,0.67)(3,4.0)(4,30.0)};\addlegendentry{CIL++}
    \addplot[color=blue] coordinates {
    (1,10.67)(2,34.0)(3,58.00)(4,68.0)}; \addlegendentry{w/$\mathcal{L}_{\text{att}}$}
    \addplot[draw=none, name path=sr_max,color=blue!70] coordinates {(1,10.67+2.31)(2,34.0+8.25)(3,58.00+4.27)(4,68.0+3.63)};
    \addplot[draw=none, name path=sr_min,color=blue!70] coordinates {(1,10.67-2.31)(2,34.0-8.25)(3,58.00-4.27)(4,68.0-3.63)};
    
    \addplot[blue!50,fill opacity=0.5] fill between[of=sr_max and sr_min];
    \addplot[draw=none, name path=cpp_max,color=red!70] coordinates {(1,0.0+0.0)(2,0.67+1.15)(3,4.0+2.0)(4,30.0+8.72)};
    \addplot[draw=none, name path=cpp_min,color=red!70] coordinates {(1,0.0-0.0)(2,0.67-1.15)(3,4.0-2.0)(4,30.0-8.72)};
    
    \addplot[red!50,fill opacity=0.5] fill between[of=cpp_max and cpp_min];

\nextgroupplot[
    xlabel={\Large Driving Weathers},
    ylabel={\Large Driving Score (DS)},
    xtick={1,2,3,4}
]
    \addplot[color=red,dashed] coordinates {(1,10.45)(2,23.06)(3,25.99)(4,52.64)};
    \addplot[color=blue] coordinates {
    (1,37.69)(2,56.79)(3,75.83)(4,76.90)}; 
    \addplot[draw=none, name path=sr_max,color=blue!70] coordinates {(1,37.69+4.23)(2,56.79+8.25)(3,75.83+4.27)(4,76.90+3.63)};
    \addplot[draw=none, name path=sr_min,color=blue!70] coordinates {(1,37.69-4.23)(2,56.79-8.25)(3,75.83-4.27)(4,76.90-3.63)};
    
    \addplot[blue!50,fill opacity=0.5] fill between[of=sr_max and sr_min];
    \addplot[draw=none, name path=cpp_max,color=red!70] coordinates {(1,10.45+0.38)(2,23.06+1.18)(3,25.99+1.13)(4,52.64+4.72)};
    \addplot[draw=none, name path=cpp_min,color=red!70] coordinates {(1,10.45-0.38)(2,23.06-1.18)(3,25.99-1.13)(4,52.64-4.72)};
    
    \addplot[red!50,fill opacity=0.5] fill between[of=cpp_max and cpp_min];

\nextgroupplot[
    xlabel={\Large Driving Weathers},
    ylabel={\Large Route Completion (RC)},
    xtick={1,2,3,4}
]

    \addplot[color=red,dashed] coordinates {(1,12.19)(2,32.50)(3,36.55)(4,66.62)};
    \addplot[color=blue] coordinates {
    (1,53.84)(2,69.23)(3,84.80)(4,84.69)}; 
    \addplot[draw=none, name path=sr_max,color=blue!70] coordinates {(1,53.84+4.0)(2,69.23+7.41)(3,84.80+5.16)(4,84.69+4.59)};
    \addplot[draw=none, name path=sr_min,color=blue!70] coordinates {(1,53.84-4.0)(2,69.23-7.41)(3,84.80-5.16)(4,84.69-4.59)};
    
    \addplot[blue!50,fill opacity=0.5] fill between[of=sr_max and sr_min];
    \addplot[draw=none, name path=cpp_max,color=red!70] coordinates {(1,12.19+0.42)(2,32.50+0.63)(3,36.55+1.24)(4,66.62+2.80)};
    \addplot[draw=none, name path=cpp_min,color=red!70] coordinates {(1,12.19-0.42)(2,32.50-0.63)(3,36.55-1.24)(4,66.62-2.80)};
    
    \addplot[red!50,fill opacity=0.5] fill between[of=cpp_max and cpp_min];

\nextgroupplot[
    xlabel={\Large Driving Weathers},
    ylabel={\Large Infraction Score (IS)},
    xtick={1,2,3,4},
    ymin=0.5,ymax=1,
    ytick={0.6,0.7,0.8,0.9,1.0}
]

    \addplot[color=red,dashed] coordinates {(1,0.92)(2,0.70)(3,0.67)(4,0.73)};
    \addplot[color=blue] coordinates {
    (1,0.67)(2,0.78)(3,0.87)(4,0.87)}; 
    \addplot[draw=none, name path=sr_max,color=blue!70] coordinates {(1,0.67+0.05)(2,0.78+0.04)(3,0.87+0.02)(4,0.87+0.01)};
    \addplot[draw=none, name path=sr_min,color=blue!70] coordinates {(1,0.67-0.05)(2,0.78-0.04)(3,0.87-0.02)(4,0.87-0.01)};
    
    \addplot[blue!50,fill opacity=0.5] fill between[of=sr_max and sr_min];
    \addplot[draw=none, name path=cpp_max,color=red!70] coordinates {(1,0.92+0.01)(2,0.70+0.03)(3,0.67+0.01)(4,0.73+0.03)};
    \addplot[draw=none, name path=cpp_min,color=red!70] coordinates {(1,0.92-0.01)(2,0.70-0.03)(3,0.67-0.01)(4,0.73-0.03)};
    
    \addplot[red!50,fill opacity=0.5] fill between[of=cpp_max and cpp_min];
    
\end{groupplot}
\end{tikzpicture}

}
\caption{Driving results by incrementally adding a weather condition to the training set (2 hours of data per weather).}
\label{fig:lowdata_town02nocrash_weathers}
\end{figure*}

While our proposed approach demonstrates superior performance in scenarios with limited data, concerns about potential generalization issues arise with a lack of diverse cases. To address this, we conduct experiments by sampling the 8-hour dataset based on the accumulation of different weather conditions to evaluate behavior against a change in domain. Note that combining all 4 kinds of weather results in the same 8-hour dataset registered in Fig. \ref{fig:lowdata_town02nocrash_meanstdev}. Fig. \ref{fig:lowdata_town02nocrash_weathers} reveals that the baseline struggles to generalize effectively even when all 4 kinds of weather are included. Although it obtains a high IS score when using one weather, this is due to the agent not moving, which can be appreciated in a low RC score. In contrast, our approach demonstrates consistent gains in all metrics as we add new weather types to the training dataset, obtaining similar results with only 2 weathers compared to the 4 weathers used in the vanilla model.

These observations emphasize the robustness of using the proposed Attention Loss when confronted with challenges such as insufficient or low-variation data. This resilience contributes to the effectiveness of our approach in scenarios with limited training samples and potential domain shifts.

\subsubsection{High data regime}

In the context of incorporating attention maps during training, two common approaches are employed: Soft Mask (SM) and Hard Mask (HM). The SM method involves adding the attention map as a fourth channel concatenated to the camera images that we pass as input to the driving model. Conversely, the HM method utilizes an element-wise product between the attention mask and the camera image, emphasizing regions indicated by the mask. Unlike our method, both of these methods require attention maps during the inference phase. To this end, we employed the U$^2$-NET model \cite{Qin:2020}, trained on the 14-hour dataset. Qualitative results showing predicted masks $\widehat{\mathcal{M}}$ from the U$^2$-NET on training data (\texttt{Town01}) and unseen data (\texttt{Town02} and \texttt{Town03}) are depicted in Fig. \ref{fig:masks}, illustrating the similarity between the predicted and ground truth masks.

Table \ref{tab:attasinput_noisy} presents results from different approaches incorporating attention masks using the 14-hour dataset. The first observation is that the inclusion of attention masks consistently improves results over the baseline, except for the SM method, which gets similar driving performance. The comparison between our approach and the HM method shows a 13\% increase in completed routes, along with a longer average driving time and fewer infractions for our approach. Following the results with noisy masks we can see the same conclusions although the performance is lower for all attention-based methods due to the included noise. Note that both SM and HM methods need to have the attention mask during testing, which we avoid by only needing it during training.

\begin{table}[]
\caption{Masks as different types of input and effect of noisy  masks (train: 14h Data \texttt{Town01}, test: \texttt{Town02}, New weathers)}
    \label{tab:attasinput_noisy}
    \begin{threeparttable}
\resizebox{\columnwidth}{!}{
\begin{tabular}{@{}lccll@{}}
\toprule
\multicolumn{1}{c}{}  & $\textbf{SR}\uparrow$          & $\textbf{DS}\uparrow$          & \multicolumn{1}{c}{$\textbf{RC}\uparrow$} & \multicolumn{1}{c}{$\textbf{IS}\uparrow$} \\ \midrule \midrule
CIL++                 & $41.33 \pm 8.08$       & $60.45 \pm 4.60$       & $73.03 \pm 4.18$                  & $0.77 \pm 0.03$                   \\
\quad w/SM      & $42.00 \pm 7.21$ & $59.29 \pm 5.49$ & $70.12 \pm 4.32$ & $0.78 \pm 0.02$             \\
\quad w/HM    & $66.00 \pm 9.17$ & $77.34 \pm 6.93$ & $84.32 \pm 5.83$ & $0.87 \pm 0.04$                                \\
\quad w/$\mathcal{L}_{\text{att}}$             & $\textbf{79.33} \pm 13.01$     & $\textbf{85.67} \pm 7.84$       & $\textbf{91.13} \pm 6.21$                  & $\textbf{0.92} \pm 0.05$                   \\ \midrule
\quad w/SM + $f(\widehat{\mathcal{M}}_{i, t})$ \tnote{a} & $35.33 \pm 7.02$ & $56.38 \pm 1.32$ & $68.38 \pm 0.58$ & $0.77 \pm 0.01$ \\
\quad w/HM + $f(\widehat{\mathcal{M}}_{i, t})$ & $66.00 \pm 7.21$ & $76.36 \pm 3.72$ & $83.46 \pm 4.48$ & $0.87 \pm 0.01$ \\
\quad w/$\mathcal{L}_{\text{att}}$ + $f(\mathcal{M}_{i, t})$ \tnote{b}         & $\textbf{71.33} \pm 6.11$ & $\textbf{80.36} \pm 6.88$ & $\textbf{89.46} \pm 3.97$ & $\textbf{0.87} \pm 0.05$                                 \\ \bottomrule
\end{tabular}}
\begin{tablenotes}
      \footnotesize
      \item [a] $f(\widehat{\mathcal{M}}_{i, t})$: Noisy predicted Masks
      \item [b] $f(\mathcal{M}_{i, t})$: Noisy Masks
    \end{tablenotes}
    \end{threeparttable}
\end{table}

To expand our experiments to scenarios with abundant data and more complex cases, we used the 55-hour dataset. This dataset introduces challenges such as multi-lanes, highways, and crossroads. Examining the results in Table \ref{tab:attasinput_55}, the gap between the baseline and our approach is reduced here, yet our approach still outperforms, completing 3\% more routes with increased average distance coverage. However, the biggest difference lies in the quality of the agent driving, where our approach achieves a higher average IS, improving from 0.5 to 0.7. These results demonstrate the effectiveness of our training method when using large datasets to enhance driving quality without the need for additional perception modules during the driving phase.

\begin{table}[]
\caption{Effect of using the attention loss in the high data regime (train: 55h Data, test: \texttt{Town05}, New weathers)}
    \label{tab:attasinput_55}
\begin{threeparttable}
\resizebox{\columnwidth}{!}{
\begin{tabular}{@{}lccll@{}}
\toprule
\multicolumn{1}{c}{}  & $\textbf{SR} \uparrow$          & $\textbf{DS}\uparrow$          & \multicolumn{1}{c}{$\textbf{RC}\uparrow$} & \multicolumn{1}{c}{$\textbf{IS}\uparrow$} \\ \midrule \midrule
CIL++                 & $70.00 \pm 5.00$ & $36.46 \pm 4.03$ & $79.69 \pm 3.84$ & $0.51 \pm 0.04$                   \\
\quad w/$\mathcal{L}_{\text{att}}$            & $\textbf{73.33} \pm 5.77$ & $\textbf{58.23} \pm 4.71$ & $\textbf{82.88} \pm 1.28$ & $\textbf{0.70} \pm 0.03$       \\ \bottomrule
\end{tabular}}
\end{threeparttable}
\end{table}

\subsection{Qualitative Results}

\begin{figure*}
\centering
\includegraphics[width=\linewidth]{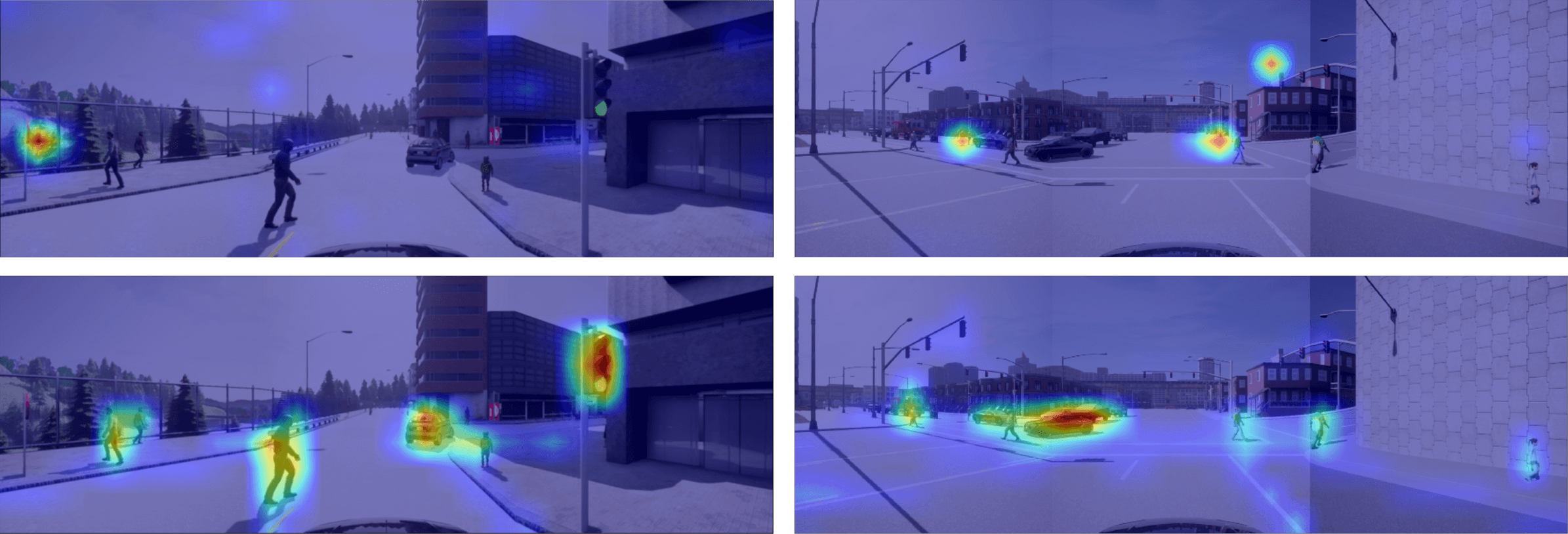}
    \caption{Visualization of the average attention map of the last layer of the Transformer Encoder using three RGB cameras as input for CIL++ (top row) and CIL++ with the Attention Loss $\mathcal{L}_\text{att}$ (bottom row), for \texttt{Town01} (left column) and \texttt{Town03} (right column).}
    \label{fig:visualization}
\end{figure*}

Visualizations of the resulting attention maps for the Transformer Encoder can be found in Fig. \ref{fig:visualization}. For the scenario in \texttt{Town01}, even though both models correctly predict to break for the incoming pedestrian, CIL++'s attention maps lack \textit{explainability}. 
Differently, thanks to the Attention Loss during training, our method provides quite \textit{explainable} and \textit{interpretable} visualizations of the attention on the sensory input. We can observe that the model has learned to segment the objects belonging to the classes of interest without needing an additional network to perform this task or to remove part of its input via masking, even in the much harder scenario in \texttt{Town03}. As a potential approach to reveal deep neural models's black-box characteristics, we encourage this work to be further explored by the community to better learn the correlation between input data and output values, beyond end-to-end driving models.

\section{Conclusions}
\label{sec:conclusions}
In this paper, we demonstrate how it is possible to guide the attention of a pure-vision end-to-end driving model by introducing a (noisy) saliency semantic map loss, without model architecture modification. Thus, no increasing computational resources are required at testing time. Using CIL++ as a reference model and the CARLA simulator with its standard benchmarks, 
we provide rich experimental results to show that our method is superior to others that require the computation of saliency maps at testing time. Our method also helps to obtain more intuitive activation maps, which we plan to use as behavior explanations in natural language. In the same vein, we plan to leverage this research to explore the field of causal correlation learning for deep learning models. Lastly, encouraged by the results using noisy attention masks, we plan to test the Attention Loss with real data and deploy the model in a real car.

\section*{Acknowledgements}
This research is supported by project TED2021-132802B-I00 funded by MCIN/AEI/10.13039/501100011033 and the European Union NextGenerationEU/PRTR. Antonio M. López acknowledges the financial support to his general research activities given by ICREA under the ICREA Academia Program. Antonio and Gabriel thank the synergies, in terms of research ideas, arising from the project PID2020-115734RB-C21 funded by MCIN/AEI/10.13039/501100011033. The authors acknowledge the support of the Generalitat de Catalunya CERCA Program and its ACCIO agency to CVC’s general activities.

{
    \small
    \bibliographystyle{ieeenat_fullname}
    \bibliography{main}
}


\end{document}